\documentclass[runningheads]{llncs}
\usepackage{graphicx}
\usepackage{comment}
\usepackage{xcolor}
\usepackage{graphicx}
\usepackage{cite}
\usepackage{amsmath,amssymb,amsfonts}
\usepackage{algorithmic}
\usepackage{textcomp}
\usepackage{xcolor}
\usepackage{comment}
\usepackage{multirow}
\usepackage{epstopdf}
\usepackage{scalerel,stackengine}
\usepackage{cite}
\usepackage{amsmath,amssymb,amsfonts}
\usepackage{algorithmic}
\usepackage{graphicx}
\usepackage{textcomp}
\usepackage{xcolor}
\usepackage{float}
\usepackage{xcolor}
\usepackage{tikz}
\usepackage{adjustbox}
\usepackage{tikz}
\usepackage{pgfplots}
\usepackage{bchart}
\usetikzlibrary{decorations.pathreplacing}
\usepackage{subfig}
\pgfplotsset{compat=1.10}
\usepackage{multibib}
\definecolor{bblue}{HTML}{4F81BD}
\definecolor{rred}{HTML}{C0504D}
\definecolor{ggreen}{HTML}{9BBB59}
\usepackage[section]{placeins}
\usepackage[figurename=Figure]{caption}
\usepackage{caption} 
\usepackage[misc,geometry]{ifsym}

\begin{document}
\title{BloomNet: Exploring Single vs. Multiple Object Annotation for Flower Recognition Using YOLO Variants}

\titlerunning{ } 

\author{Safwat Nusrat\inst{1}\Letter\orcidID{0009-0002-4037-4325} \and
Prithwiraj Bhattacharjee\inst{1}\orcidID{0000-0001-9300-9351}}

\authorrunning{ } 

\institute{Department of Computer Science and Engineering,\newline Leading University, Sylhet-3112, Bangladesh \newline
\email{\{cse\_0182210012101213, prithwiraj\_cse\}@lus.ac.bd}}

\maketitle 

%

%
%
%
\begin{abstract}
Precise localization and recognition of flowers are crucial for advancing automated agriculture, particularly in plant phenotyping, crop estimation, and yield monitoring. This paper benchmarks several YOLO architectures, such as, YOLOv5s, YOLOv8n/s/m, and YOLOv12n, for flower object detection under two annotation regimes: single-image single-bounding box (SISBB) and single-image multiple-bounding box (SIMBB). The FloralSix dataset, comprising 2,816 high-resolution photos of six different flower species, is also introduced. It is annotated for both dense (clustered) and sparse (isolated) scenarios. The models were evaluated using Precision, Recall, and Mean Average Precision (mAP) at IoU thresholds of 0.5 (mAP@0.5) and 0.5–0.95 (mAP@0.5:0.95). In SISBB, YOLOv8m (SGD) achieved the best results with Precision 0.956, Recall 0.951, mAP@0.5 0.978, and mAP@0.5:0.95 0.865, illustrating strong accuracy in detecting isolated flowers. With mAP@0.5 0.934 and mAP@0.5:0.95 0.752, YOLOv12n (SGD) outperformed the more complicated SIMBB scenario, proving robustness in dense, multi-object detection. Results show how annotation density, IoU thresholds, and model size interact: recall-optimized models perform better in crowded environments, whereas precision-oriented models perform best in sparse scenarios. In both cases, the Stochastic Gradient Descent (SGD) optimizer continuously performed better than alternatives. These density-sensitive sensors are helpful for non-destructive crop analysis, growth tracking, robotic pollination, and stress evaluation.

\end{abstract}
\section{Introduction}
\noindent The growing need for automation in ecological monitoring, horticulture, and precision farming has created more interest in computer vision for flower recognition. Flowers, and their diverse forms, scales, and colors, are key indicators of ecosystem health and biodiversity. Flowers have variations in orientation, illumination, and background clutter, which makes their detection challenging in real-world scenes \cite{Wang2025}. When multiple flowers appear in the same frame, sometimes overlapping, partially hidden, or forming dense clusters, traditional single-label classification methods become inadequate, especially in natural or cultivated settings.
Existing datasets, such as Oxford102, Oxford17, and Iris dataset, primarily focus on classification and offer limited or inconsistent bounding box annotations, which restrict their use in analyzing object density or performing robust multi-flower detection. Driven by these shortcomings, this paper presents a fully labeled flower dataset and argues about the performances of the state-of-the-art deep architectures of the YOLOs family, namely, the YOLOv5, the YOLOv8, and the YOLOv12 models, to achieve real-time representation of multiple flowers within one picture, with the impact of the object density on the model performance analyzed.
\textbf{This work is novel in several ways:} 
\begin{itemize}
    \item Roboflow \cite{Roboflow2025} tool-based labels of annotated FloralSix \footnote{\url{https://www.kaggle.com/datasets/arefin07/6-class-flower-dataset}} dataset, with bounding boxes for each class of flowers to detect flower objects. 
    \item The first ever establishment of an experimental baseline for flower object detection on this dataset.
    \item Detection of all flower variants in a single image, while the traditional classification models can deal with only one variant of flower per image.
    \item Inclusion of the object density analysis, which analyzes the performance of the detection between the sparse and the dense flower distribution.
\end{itemize}

\noindent This work develops a baseline evaluation framework, dense YOLO-based analysis, and thorough annotation to improve flower detection. It promotes upcoming advancements in large-scale monitoring and model creation while filling a significant gap in the literature. The findings show how deep learning models may be used practically in floral environments, which has ramifications for environmental, horticultural, and ecological applications.
\section{Related Work}
For automated harvesting, pollination monitoring, and yield prediction, flower detection is essential to precision agriculture. Recent CNN-based deep learning models, like YOLO, effectively handle challenges like dense clustering, changeable illumination, and occlusion. \cite{Guo2025,Wang2025-Yo_AFD,plants14030468,Ang2025,Wang2025,Wu2024,horticulturae10050517,Stark2023,Cheng2020,Lu2024,Bhattarai2024}.
A primary focus has been made on the improvement of YOLO architecture to work in real time. MSDP-Net Guo et al. \cite{Guo2025} presented MSDP-Net, a YOLOv5m variant with Convolutional Block Attention Module (CBAM) and dynamic viewpoint adaptation using binocular cameras, trained on 6400 augmented safflower images, raising the positioning accuracy to 93.79\% and the mAP score by 5.5\%compared to base. 
The proposed architecture of Wang et al. \cite{Wang2025-Yo_AFD} was YO-AFD based on YOLOv8 with ISAT, C2f-IS feature fusion, and Focaler IoU loss, trained on 2115 apple flower photos, outperforming YOLOv5n/YOLOv8n.
Uniformly, Wang et al. \cite{plants14030468} developed VM-YOLO, augmenting YOLOv8 with Light C2f and VMambaNeck for strawberry detection. Using 3388 augmented plantation images, it produces  71.4\% mAP, fewer parameters (30M), and faster processing. 
Wang and Zhang \cite{Wang2025} developed an improved YOLOv8 model to perfectly detect chili pepper flowers, essential for robotic pollination in greenhouses. The model combines a C-ELAN module 
and combines a GMBAM module to highlight object edges. 
For UAV-based apricot flower detection, Wu et al. \cite{Wu2024} modified D-YOLOv8 with D-FPN and DAL, obtaining 99.38\% AP@0.5 and 123 FPS on 3600 complex and 870 low-light images (257k instances). 
Going further than pure YOLO, Li et al. \cite{horticulturae10050517} had added masked autoencoders (MAEs) (combined with Vision Transformer (ViT-Large) and the YOLOv5), to recognize and detect flowers. 
Stark et al.\cite{Stark2023} utilized YOLOv5n, YOLOv5s and YOLOv7t for detecting flower-visiting arthropods, facilitating ecological studies. With 17,708 GBIF images across eight groups, it reached 96.24\% accuracy and >78\% IoU, endowing real-time on low-end hardware.
With hybrid connections and 16× downsampling, Lu et al. \cite{Lu2024} MAR-YOLOv9 improves YOLOv9 for rapeseed flower detection. It achieves 9.3\% smaller size and 1.8\% higher mAP@0.5 on RFRB, although it is still impacted by occlusion and variations in lighting.
AgRegNet is a U-Net-like regression network proposed by Bhattarai et al.\cite{Bhattarai2024} and ConvNeXt-T on apples flowers density and localization. It outperformed CSRNet (18.1 MAE) with 0.938 SSIM, 13.7\% pMAE, and 0.81 mAP on 325 orchard images.
Among the gaps are multi-species generalization, where most studies \cite{Guo2025, Wang2025-Yo_AFD,plants14030468},\cite{Wang2025}, \cite{Wu2024}, \cite{Bhattarai2024} base their work on single-flower-species datasets, and this may miss potential inter-species variations. Although \cite{horticulturae10050517} uses the widely used multi-class Oxford 102. Future studies should concentrate on different, multispecies standardized databases and hybrid detectors with reasoning functions.
\section{Methodology}

\noindent This section outlines the full methodology of the current research to be used in order to assess the impact of object density on several versions of YOLO (You Only Look Once) when conducting flower-detection. The methodology includes the preparation of datasets, the choice of the model, network structure, object density processing, loss functions, metrics, and optimization. The overall framework is designed to systematically analyze the performance of object detection models under varying conditions of floral density.

\subsection{Dataset Preparation}

\noindent Our research utilized the FloralSix dataset, sourced from Kaggle \cite{Safwat2025}. This dataset comprises 2,816 high-resolution images collected from various gardens in Bangladesh. Hibiscus (225), marigold (649), zinnia (672), melampodium (582), crape jasmine (135), and Madagascar periwinkle (553) are among the six flower species that are included. Because the photos were taken outside, they present a range of backgrounds, lighting conditions, and angles that present practical difficulties for computer vision tasks. The images are in JPG format. There are no set sizes, but the resolution is primarily set to 3000x4000 pixels. No prior object detection study has ever used this data. This is what distinguishes and elevates our work.

\subsubsection{Data Annotation and Preprocessing}

\noindent The FloralSix\footnote{\url{https://www.kaggle.com/datasets/safwatnusrat/floralsix-annotation}} dataset was annotated in two stages using Roboflow \cite{Roboflow2025} to handle varying object densities and image complexities. In order to establish a localization baseline, each image (a total of 2816) was annotated with a single bounding box that highlighted the main flower. In order to investigate the effect of object density on detection performance, the second phase used a thorough approach by annotating several bounding boxes per image. As indicated in Table~\ref{datasetComparison}, this procedure produced 6,934 bounding boxes in the six flower-type images, with several instances having two to twenty-six bounding boxes. FloralSix is positioned as a benchmark for flower object detection thanks to this custom dataset preparation, especially in multi-instance scenarios that are frequently disregarded in current floral datasets. For research purposes, the dataset is publicly accessible on Kaggle\cite{Annotation}.
\vspace{-10mm}
\begin{table}[h!]
\centering
\caption{Annotation statistics of the FloralSix dataset.}
\label{datasetComparison}
\begin{tabular}{|c|c|c|}
\hline
\textbf{Annotation Type} & \textbf{No. of Images} & \textbf{No. of Objects} \\
\hline
Single Bounding Box & 2816 & 2816 \\
\hline
Multiple Bounding Box & 2816 & 6934 \\
\hline
\end{tabular}
\end{table}
\vspace{-15mm}
\begin{table}[h!]
\centering
\caption{Comparison of training, validation, and test dataset annotations.}
\label{datasetComparison of train test split}
\begin{tabular}{|c|c|c|c|c|c|c|}
\hline
\multirow{2}{*}{\textbf{Annotation Type}} & \multicolumn{3}{c|}{\textbf{No. of Images}} & \multicolumn{3}{c|}{\textbf{No. of Objects}} \\
\cline{2-7}
 & \textbf{Train} & \textbf{Valid} & \textbf{Test} & \textbf{Train} & \textbf{Valid} & \textbf{Test} \\
\hline
Single Bounding Box & 1972 & 422 & 422 & 1972 & 422 & 422 \\
\hline
Multiple Bounding Box & 1972 & 422 & 422 & 4905 & 917 & 1112 \\
\hline
\end{tabular}
\end{table}

\subsubsection{Dataset Format and Structure}

\noindent The dataset is formatted for object detection, adhering to the YOLO standard. All images were preprocessed by being converted to the same size of 640x640 pixels to facilitate efficient input and stability of the model. Table~\ref{datasetComparison of train test split} shows that the data is split into three groups: 75\% for training, 15\% for testing, and 15\% for validation. Each subset contains two primary subdirectories:

\begin{itemize}
    \item \textbf{images/}: This is the directory where the image files (e.g. .jpg, .png) of the corresponding subset are stored.
    \item \textbf{labels/}: These label files are plain text (.txt) files naming one per image and of the same base filename (e.g., image1.jpg has the corresponding label image1.txt). Each line in a label file represents a bounding box annotation in YOLO format: \textit{class\_id center\_x center\_y width height}. The coordinates are normalized to the range [0, 1] relative to the image size.
\end{itemize}

\subsection{Model Selection and Network Overview}

\begin{figure}[h!]
\centering
  \includegraphics[width=\linewidth]{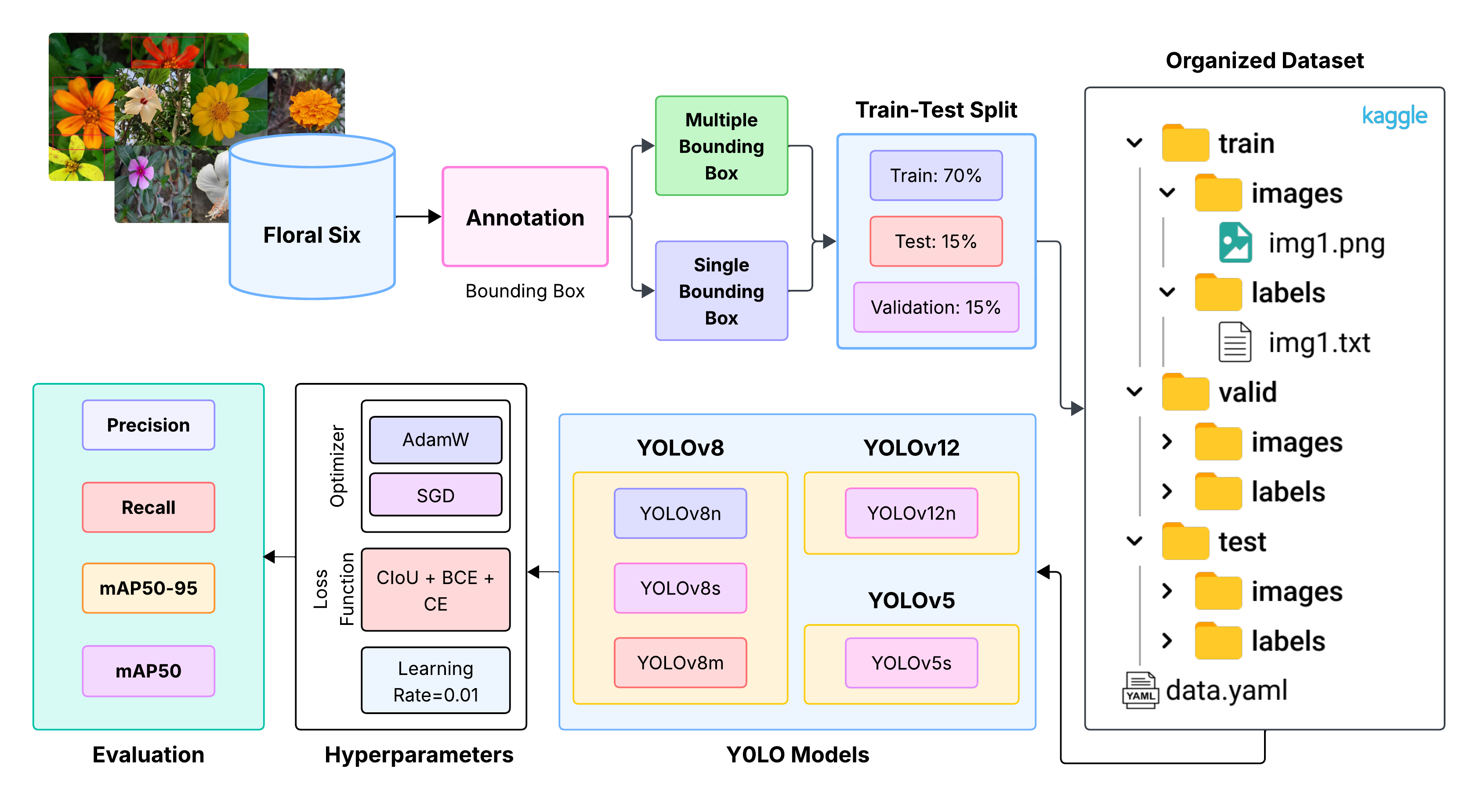}
  \caption{Model architecture for flower detection using YOLO variants.}
  \label{fig:model_architecture}
\end{figure}

\noindent Five YOLO model variations are used in this study: \textbf{YOLOv5s}, \textbf{YOLOv8n\slash s\slash m}, and \textbf{YOLOv12s}, as shown in Figure~\ref {fig:model_architecture}. YOLOv5s is a light-out baseline that is designed for low-resource deployment and speed. By integrating improved necks and CSP backbones, the YOLOv8 variant (nano, small, and medium) increases the detection of tiny or overlapping flowers. YOLOv12 uses anchor-free prediction techniques and is designed for small objects and densely populated environments. The efficacy across various flower densities was assessed using both the \textbf{SISBB} (single flower) and \textbf{SIMBB} (all flowers) methods.

\subsubsection{Network Architecture}
\noindent An end-to-end architecture characterizes the single-stage YOLO models. Effective real-time flower recognition is accelerated by it. The network is made up of three primary parts:

\begin{itemize}
    \item The hierarchical features are extracted using convolutional layers. For small or overlapping objects, the combination of YOLOv8 and YOLOv12's CSP (Cross Stage Partial) modules serves as an efficient way to promote gradient flow and feature reuse.
    \item Multi-scale features are accumulated in the neck. It makes use of architectures like PANet (Path Aggregation Network) and FPN (Feature Pyramid Network). Deep semantic features are combined with superficial and high-resolution features in this process. As a result, it guarantees accurate detection in a variety of floral sizes and densities.
    \item \textbf{Head}: The head has produced bounding-box coordinates, objectness scores, and class probabilities. Non-Maximum Suppression (NMS) is used to stop redundant detection in a scene with a high population density. It ensures precise localization in this way.

\end{itemize}
\noindent The suggested architecture strikes the best possible balance between accuracy and speed. Both sparse and dense flower environments can benefit from it.
\subsection{Handling Object Density}
\noindent Object density was categorized into dense (closely packed and potentially overlapping flowers) and sparse (separated flowers) classes. Two different annotation strategies were used during training. They are SIMBB (all visible flowers labeled) and SISBB (one prominent flower labeled per image). SIMBB enhances annotation complexity, proving advantageous for detection in packed scenarios. In contrast, SISBB simplifies the learning process but is less effective in dense scenes due to reduced annotation complexity. 

\subsection{Hyperparameters and Evaluation Metrics}
\noindent The loss function was divided into three components. These include classification loss, objectness loss, and bounding box loss. The bounding box loss is based on the Complete Intersection over Union (CIoU), which takes aspect ratio, distance, and overlap into account to guarantee more precise localization. While objectness loss determines whether an object is included in a projected box, classification loss guarantees that the appropriate category is assigned.
To measure performance, YOLO employs standard object detection evaluation metrics. The basis for detection accuracy is Intersection over Union (IoU), which measures the overlap between predicted and ground-truth boxes. Precision shows the percentage of accurate positive predictions, whereas recall counts the number of real items found. Overall performance is provided by the mean across classes (mAP), whereas average precision (AP) summarises precision–recall trade-offs per class. Results are typically reported at IoU thresholds such as mAP@0.5 and mAP@0.5:0.95.

\subsection{Training Strategy}
\noindent The training was carried out in the GPU-enabled environment, maintaining consistent hyperparameters across all YOLO models. To allow uniform feature extraction across both the size and density of the flowers, resizing was done to an optimal dimension of 640 x 640 pixels by considering a batch size of 16 images. The parameters of the model were optimized using SGD and AdamW optimizers. The initial learning rate was 0.01. The model ran 100 epochs in training. If the mAP@0.5 measure did not improve over 10 consecutive epochs, it utilizes an early stopping mechanism that halted the process. At the end of every epoch, the relative accuracy, recall, and mAP were validated to exhibit effective generalization.

\section{Result Analysis}
\noindent The effectiveness of different YOLO models, namely YOLOv5s, YOLOv8n/s/m, and YOLOv12n, has been demonstrated under two different annotation strategies: Single Image Multiple Bounding Box (SIMBB) and Single Image Single Bounding Box (SISBB). The main assessment is the mean Average Precision (mAP50-95) at IoU thresholds between 0.5 and 0.95. It provides a reliable indicator of object localization and classification precision. All models and optimizers' performance metrics are shown in Table~\ref{yolo_metrics}.
\vspace{-6mm}
\begin{table}[h!]
\centering
\caption{Performance metrics of different YOLO models.}
\label{yolo_metrics}
\begin{tabular}{|l|l|l|c|c|c|c|}
\hline
\textbf{Annotation Type} & \textbf{Model} & \textbf{Optimizer} & \textbf{Precision} & \textbf{Recall} & \textbf{mAP50} & \textbf{mAP50-95} \\
\hline
\multirow{10}{*}{SISBB} 
& \multirow{2}{*}{YOLOv5s} & SGD   & 0.899& 0.952 & 0.968& 0.814 \\
&                           & AdamW & 0.864 & 0.913 & 0.943 & 0.764 \\
\cline{2-7}
& \multirow{2}{*}{YOLOv8n} & SGD   & 0.96 & 0.944 & 0.978 & 0.849 \\
&                           & AdamW & 0.926 & 0.941 & 0.966 & 0.837 \\
\cline{2-7}
& \multirow{2}{*}{YOLOv8s} & SGD   & 0.931 & 0.971 & 0.979 & 0.859\\
&                           & AdamW & 0.954 & 0.956 & 0.978 & 0.847 \\
\cline{2-7}
& \multirow{2}{*}{YOLOv8m}  & SGD   & 0.956 & 0.951 & 0.978 & 0.865 \\
&                           & AdamW & 0.923 & 0.927 & 0.963 & 0.812 \\
\cline{2-7}
& \multirow{2}{*}{YOLOv12n} & SGD   & 0.941 & 0.956 & 0.979 & 0.857 \\
&                           & AdamW & 0.929& 0.88 & 0.942 & 0.8 \\
\hline
\multirow{10}{*}{SIMBB} 
& \multirow{2}{*}{YOLOv5s} & SGD   & 0.912 & 0.838 & 0.92 & 0.697 \\
&                           & AdamW & 0.884 & 0.844 & 0.893 & 0.663 \\
\cline{2-7}
& \multirow{2}{*}{YOLOv8n} & SGD   & 0.902 & 0.864 & 0.929 & 0.737 \\
&                           & AdamW & 0.927 & 0.767 & 0.882 & 0.671 \\
\cline{2-7} 
& \multirow{2}{*}{YOLOv8s} & SGD   & 0.906 & 0.864 & 0.927 & 0.725 \\
&                           & AdamW & 0.872 & 0.844 & 0.907 & 0.697 \\
\cline{2-7}
& \multirow{2}{*}{YOLO8m}  & SGD   & 0.916 & 0.85 & 0.921 & 0.726 \\
&                           & AdamW & 0.913 & 0.818 & 0.916 & 0.71 \\
\cline{2-7} 
& \multirow{2}{*}{YOLOv12n} & SGD   & 0.924 & 0.869 & 0.934 & 0.752 \\
&                           & AdamW & 0.913 & 0.842 & 0.911 & 0.713 \\
\hline
\end{tabular}
\end{table}
\subsection{Performance Under SISBB and SIMBB Annotation Strategies}
\noindent In the SISBB case, all the evaluated models were very precise, with only one labeled object in every image. YOLOv8m was the top model with a precision of 0.956, a recall of 0.951, a mAP50 of 0.978, and a mAP50-95 of 0.865, which was trained by using the SGD optimizer. These results show its ability to distinguish and best position a single object within an image. In single-object identification tasks, the SGD optimizer always achieved better results than AdamW on the majority of models, which means more stable and reliable convergence. Moreover, the models and, in particular, YOLOv8m, also show excellent results in the identification process in environments with individual, single objects. The case of SIMBB, where a few labeled objects are present in a single picture, is bound to be even more complex due to the increased density of objects and potential occlusions. As a result, the overall performance of all models was lower than in the case of SISBB. However, YOLOv12n, which uses the SGD optimizer, proved to be the lowest performer in this difficult setting with the best mAP50 of 0.934 and mAP50-95 of 0.752. This brings out the improved generalization and ability of YOLOv12n to deal with multi-object detection. AdamW-based YOLOv8 showed lower recall in SIMBB but retained precision, as well as the significance of the optimizer in multi-objective detection.



\subsection{Comparative Analysis and Key Findings}
\begin{figure}[h!]
\centering
  \includegraphics[width=\linewidth]{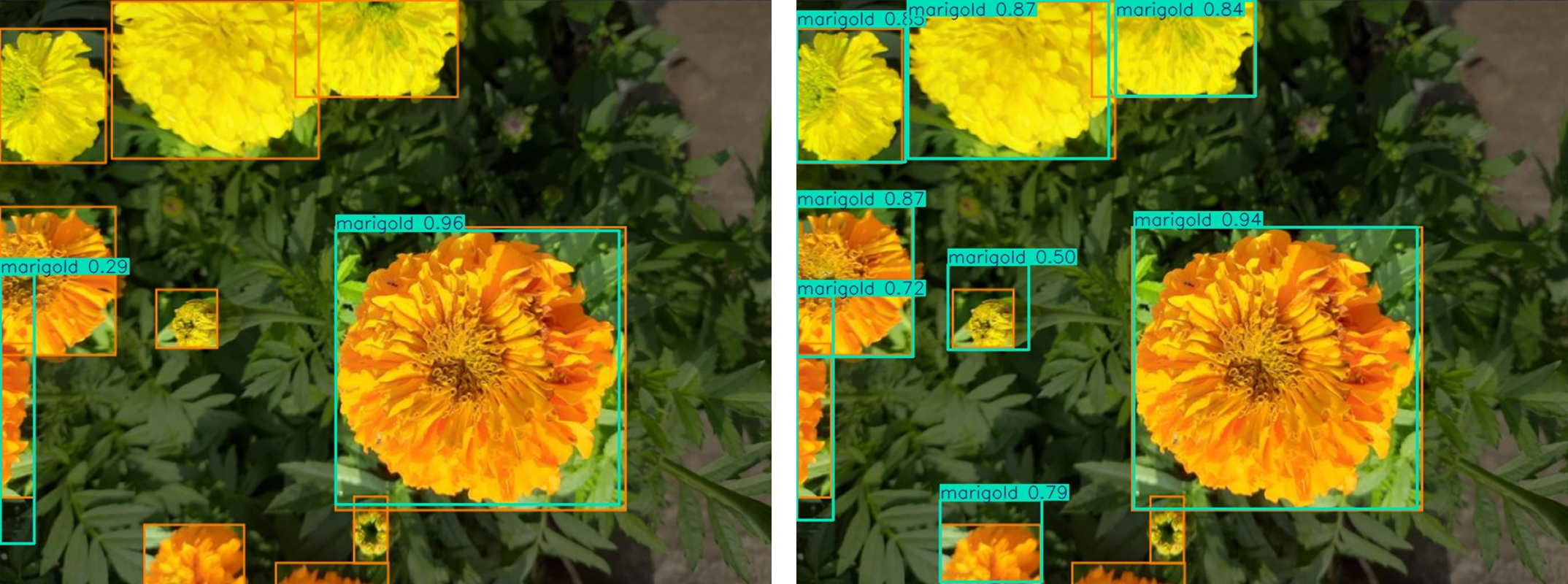}
  \caption{Comparison of object detection results under different annotation strategies: (a) SISBB (Single Image Single Bounding Box) and (b) SIMBB (Single Image Multiple Bounding Boxes). Ground-truth boxes are shown in orange, while predicted boxes are shown in blue.}
 \label{fig:comparison_image}
\end{figure}
\noindent 
Our findings show that the two annotation methodologies evidently trade off with each other. The SISBB approach achieves higher precision and mAP, making it effective for detecting single prominent objects. Despite having lower overall metrics, SIMBB provides valuable insights into a model's ability to work with complex, dense scenes. It is noteworthy that models like YOLOv12n show a strong potential to identify multiple flowers in one image, which is crucial when working in the real world, where the objects are often placed in the same place or grouped together.
In particular, to identify single and salient flowers, the SISBB strategy with such models as YOLOv8m (SGD) is more effective. Nevertheless, with the new goal to detect every instance of flowers in a congested environment, the SIMBB strategy, especially using YOLOv12n (SGD) becomes more effective. This can be visualized in Figure~\ref{fig:comparison_image}, where the difference in the detection of two methods is shown.
Overall, YOLOv8 and YOLOv12n generally outperform YOLOv5 across both scenarios. In the YOLOv8 series, scaling the models from nano (n) to small (s) and medium (m) variations usually results in improved accuracy, albeit at the expense of larger computing demands. The relevance of the SGD optimizer to these object detection tasks was also enabled by the fact that it always performed better than AdamW in both types of annotations. In single- and multi-object detection, YOLOv12n is one of the most balanced solutions that offer a decent combination of consistency, accuracy, and scalability. For efficient deployment, YOLOv8m (49.6 MB) is well-suited for SISBB, while the smaller YOLOv12n (5.6 MB) provides faster inference for SIMBB. In real-time agricultural applications, this depicts the trade-off between accuracy and model size.

\section{Conclusion}
\noindent This research evaluates the performance of different YOLO architectures for detecting flowers at different object densities using a single-image multiple-bound box (SIMBB) and a single-image single-bound box (SISBB) annotation. By carefully preparing datasets, choosing models, and analyzing metrics, we evaluated their suitability for agricultural applications. Annotation density has a substantial effect on performance, as the results show.  The highest precision and mAP for low-density detection (SISBB) were obtained by YOLOv8m with SGD when identifying single flowers. YOLOv12n with SGD demonstrated excellent recall and generalization for overlapping cases in high-density environments (SIMBB). SGD continuously outperformed AdamW in both cases, demonstrating its dependability for YOLO training. The assessed YOLO variants can be used on UAVs or autonomous field robots for real-time flower monitoring, depending on trade-offs in model size, computation, and accuracy. As a result, it can make precision farming applications like crop stress assessment, growth stage monitoring, and robotic pollination scalable and automated.


%
%
%
%

\bibliographystyle{splncs04}
\bibliography{Bibliography}
\end{document}